\documentclass{sig-alternate-2013}
\usepackage{tabularx,booktabs}
\usepackage{colortbl}
\usepackage{algorithm}
\usepackage{algorithmic}
\usepackage{float}
\usepackage{xcolor}
\usepackage{multirow}
\usepackage{booktabs}
\usepackage{makecell}
\newfloat{algorithm}{h}{lop}
\usepackage{array}

\usepackage{subfigure}
\usepackage{xspace}
\usepackage[show]{chato-notes}

\newfont{\mycrnotice}{ptmr8t at 7pt}
\newfont{\myconfname}{ptmri8t at 7pt}
\let\crnotice\mycrnotice%
\let\confname\myconfname%

\permission{Permission to make digital or hard copies of all or part of this work for personal or classroom use is granted without fee provided that copies are not made or distributed for profit or commercial advantage and that copies bear this notice and the full citation on the first page. Copyrights for components of this work owned by others than ACM must be honored. Abstracting with credit is permitted. To copy otherwise, or republish, to post on servers or to redistribute to lists, requires prior specific permission and/or a fee. Request permissions from Permissions@acm.org.}
\conferenceinfo{CIKM'15,}{October 19--23, 2015, Melbourne, Australia.} 
\copyrightetc{\copyright~2015 ACM. ISBN \the\acmcopyr}
\crdata{978-1-4503-3794-6/15/10\ ...\$15.00.\\
DOI: http://dx.doi.org/10.1145/XXX.XXXXXXX}

\begin{document}
\title{A Hierarchical Recurrent Encoder-Decoder\\
for Generative Context-Aware Query Suggestion}
\numberofauthors{6}
\author{
Alessandro Sordoni$^{f\ddag}$, \;
Yoshua Bengio$^f$, \;
Hossein Vahabi$^g$, \;
\and  
Christina Lioma$^h$, \;
Jakob G. Simonsen$^h$, \;
Jian-Yun Nie$^f$ \vspace{1mm}\\
\affaddr{$^f$DIRO, Universit\'e de Montr\'eal, Qu\'ebec}\\
\affaddr{$^g$Yahoo Labs, Barcelona, Spain}\\
\affaddr{$^h$Dept. Computer Science, Copenhagen University, Denmark}\vspace{1mm} \\
$^\ddag$\email{sordonia@iro.umontreal.ca}}
\maketitle
\begin{abstract}
Users may strive to formulate an adequate textual query for their information need. 
Search engines assist the users by presenting query suggestions.
To preserve the original search intent, suggestions should be context-aware and account for the previous queries issued by the user.
Achieving context awareness is challenging due to data sparsity.
We present a probabilistic suggestion model that is able to account for sequences of previous queries of arbitrary lengths.
Our novel hierarchical recurrent encoder-decoder architecture allows the model to be sensitive to the order of queries in the context while avoiding data sparsity.
Additionally, our model can suggest for rare, or long-tail, queries.
The produced suggestions are synthetic and are sampled one word at a time, using computationally cheap decoding techniques.
This is in contrast to current synthetic suggestion models relying upon machine learning pipelines and hand-engineered feature sets.
Results show that it outperforms existing context-aware approaches in a next query prediction setting.
In addition to query suggestion, our model is general enough to be used in a variety of other applications.
\end{abstract}
%

\vspace{1mm}
\noindent
{\bf Categories and Subject Descriptors:} H.3.3 {[Information Search and Retrieval]}: {Query formulation}

\vspace{2mm}
\noindent
{\bf Keywords:} Recurrent Neural Networks; Query Suggestion.
\begin{figure*}[t]
\centering
\includegraphics[scale=0.20]{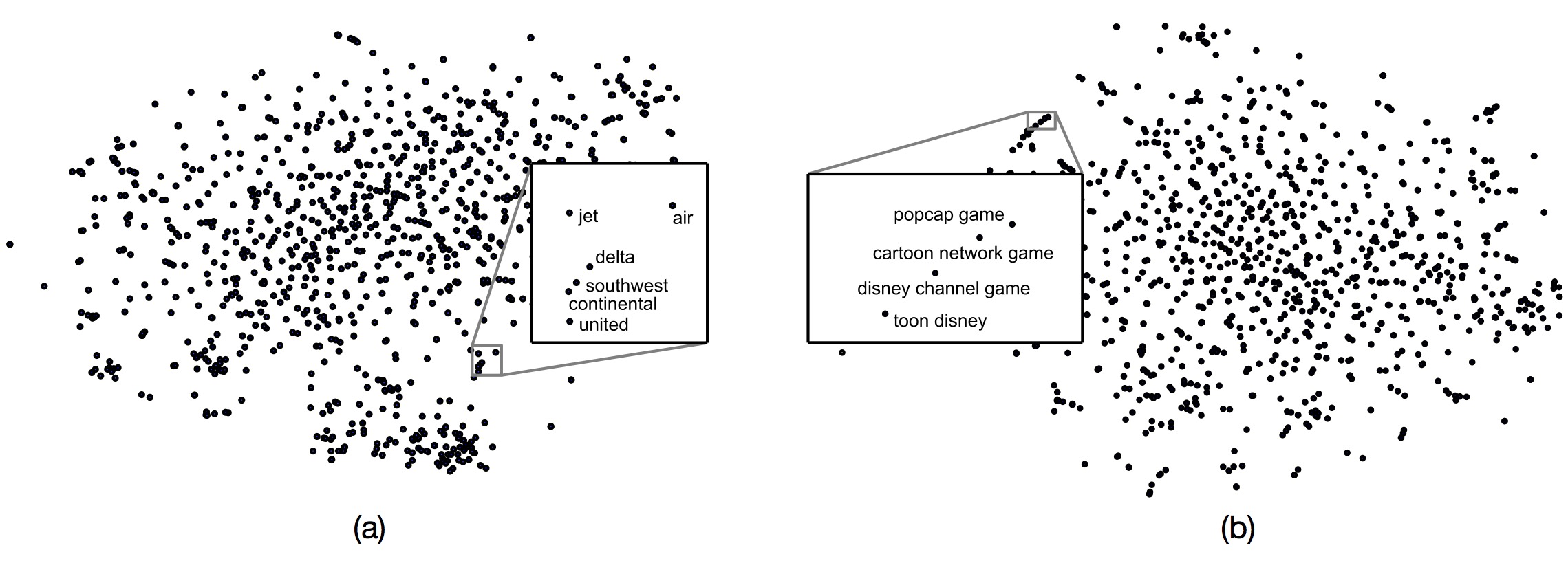}
\caption{\label{fig:word_query_embeddings}Projection of the (a) word and (b) query embeddings learnt by our neural network architecture. Topically similar terms and queries are close in the embedding space.}
\end{figure*}

\section{Introduction}
\label{sec:intro}

Modern search engines heavily rely on query suggestions to support users during their search task.
Query suggestions can be in the form of auto-completions or query reformulations.
Auto-completion suggestions help users to complete their queries while they are typing in the search box.
In this paper, we focus on query reformulation suggestions, that are produced after one or more queries have already been submitted to the search engine. 

Search query logs are an important resource to mine user reformulation behaviour. The query log is partitioned into query sessions, i.e.~sequences of queries issued by a unique user and submitted within a short time interval.  A query session contains the sequence of query reformulations issued by the user while attempting to complete the search mission. Therefore, query co-occurrence in the same session is a strong signal of query relatedness and can be straightforwardly used to produce suggestions.

Methods solely relying on query co-occurrence are prone to data sparsity and lack coverage for rare and \emph{long-tail} queries, i.e.~unseen in the training data.
A suggestion system should be able to translate infrequent queries to more common and effective formulations based on similar queries that have been seen in the training data. 
Amongst the interesting models that have been proposed, some capture higher order collocations~\cite{Boldi2008}, consider additional resources~\cite{Jain2011,Vahabi2013}, move towards a word-level representation~\cite{Bonchi2012,Broccolo2012} or describe queries using a rich feature space and apply learning to rank techniques to select meaningful candidates~\cite{Ozertem2012,Santos2013}. 

An additional desirable property of a suggestion system is \emph{context-awareness}. 
Pairwise suggestion systems operate by considering only the most recent query.
However, previous submitted queries provide useful context to narrow down ambiguity in the current query and to produce more focused suggestions~\cite{Jiang2014}. 
Equally important is the order in which past queries are submitted, as it denotes generalization or specification reformulation patterns~\cite{Huang2009}. A major hurdle for current context-aware models is dealing with the dramatic growth of diverse contexts, since it induces sparsity, and classical count-based models become unreliable~\cite{Cao2008,He2009}.


Finally, relatively unexplored for suggestion systems is the ability to produce \emph{synthetic} suggestions. 
Typically, we assume that useful suggestions are already present in the training data. 
The assumption weakens for rare queries or complex information needs, for which it is possible that the best suggestion has not been previously seen~\cite{Jain2011, Templates11}. 
In these cases, synthetic suggestions can be leveraged to increase coverage and can be used as candidates in complex learning to rank models~\cite{Ozertem2012}.

We present a generative probabilistic model capable of producing synthetic, context-aware suggestions not only for popular queries, but also for long tail queries.
Given a sequence of queries as prefix, it predicts the most likely sequence of words that follow the prefix. 
Variable context lengths can be accounted for without strict built-in limits.
Query suggestions can be mined by sampling likely continuations given one or more queries as context.
Prediction is efficient and can be performed using standard natural language processing word-level decoding techniques~\cite{Koehn2009}.
The model is robust to long-tail effects as the prefix is considered as a sequence of words that share statistical weight and not as a sequence of atomic queries.

As an example, given a user query session composed of two queries \emph{cleveland gallery} $\rightarrow$ \emph{lake erie art} issued sequentially, our model predicts sequentially the words \emph{cleveland}, \emph{indian}, \emph{art} and $\circ$, where $\circ$ is a special end-of-query symbol that we artificially add to our vocabulary.
As the end-of-query token has been reached, the suggestion given by our model is \emph{cleveland indian art}.
The suggestion is contextual as the concept of \emph{cleveland} is justified by the first query thus the model does not merely rely on the most recent query only.
Additionally, the produced suggestion is synthetic as it does not need to exist in the training set.

To endow our model with such capabilities, we rely on recent advances in generative natural language applications with neural networks~\cite{Bengio2013,Kyung14,Bhaskar15}.
We contribute with a new hierarchical neural network architecture that allows to embed a complex distribution over sequences of queries within a compact parameter space. 
Differently from count-based models, we avoid data sparsity by assigning single words, queries and sequences of queries to embeddings, i.e.~dense vectors bearing syntactic and semantic characteristics (Figure~\ref{fig:word_query_embeddings})~\cite{bengio2003neural}.
Our model is compact in memory and can be trained end-to-end on query sessions.
We envision future applications to various tasks, such as search log mining, query auto-completion and query next-word prediction.


\section{Key Idea}
\label{sec:overview}
Suggestion models need to capture the underlying similarities between queries.
Vector representations of words and phrases, also known as \emph{embeddings}, have been successfully used to encode syntactic or semantic characteristics thereof~\cite{Bengio2013,bengio2003neural,mikolov2013efficient,shen2014latent}. 
We focus on how to capture query similarity and query term similarity by means of such embeddings.
In Figure~\ref{fig:word_query_embeddings}~(a)~and~(b), we plot a two-dimensional projection of the word and query embeddings learnt by our model.
The vectors of topically similar terms or queries are close to each other in the vector space.

\begin{figure}[t]
\centering
\includegraphics[scale=0.6]{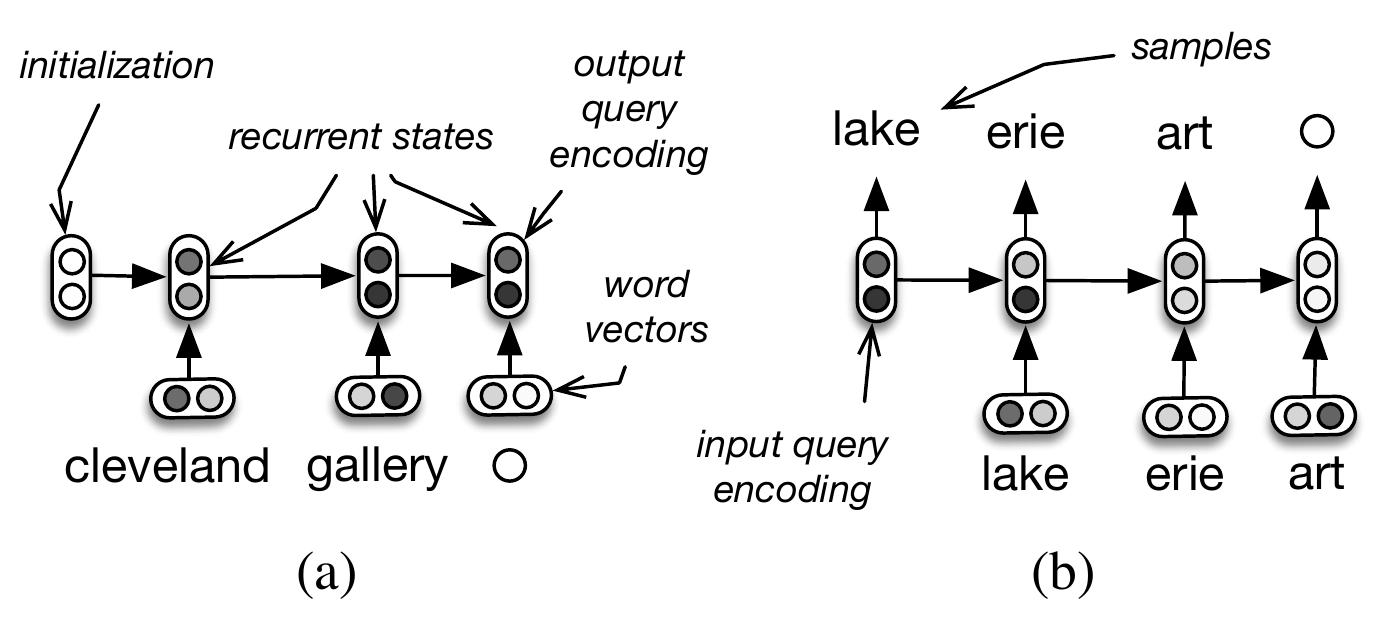}
\caption{\label{fig:rnn_intro}  (a) An encoder RNN processing the 
query \emph{cleveland gallery} followed by a special
end-of-query symbol $\circ$. Each solid arrow represents a non-linear transformation.
(b) A decoder RNN generating the next query in the session,
\emph{lake erie art}, from a query encoding as input.}
\end{figure}

Vector representations for phrases can be obtained by averaging word vectors~\cite{mikolov2013efficient}.
However, the order of terms in queries is usually important~\cite{sordoni2013modeling}.
To obtain an order-sensitive representation of a query, we use a particular neural network architecture called Recurrent Neural Network  (RNN)~\cite{Bengio2013,Mikolov2010}.
For each word in the query, the RNN takes as input its embedding and updates an internal vector, called \emph{recurrent} state,
that can be viewed as an order-sensitive summary of all the information seen up to that word. 
The first recurrent state is usually set to the zero vector.
After the last word has been processed, the recurrent state can be considered as a compact order-sensitive encoding of the query (Figure~\ref{fig:rnn_intro}~(a)).

A RNN can also be trained to decode a sentence out of a given query encoding.
Precisely, it parameterizes a conditional probability distribution on the space of possible queries given
the input encoding.
The process is illustrated in Figure~\ref{fig:rnn_intro}~(b).
The input encoding may be used as initialization of the recurrence.
Then, each of the recurrent states is used to estimate the probability of the next word in the sequence. 
When a word is sampled, the recurrent state is updated to take into account the generated
word. The process continues until the end-of-query symbol $\circ$ is produced.

The previous two use cases of RNNs can be pipelined into a single recurrent encoder-decoder, as proposed in~\cite{Kyung14,Ilya14} for Machine Translation purposes. The architecture can be used to parameterize a mapping between sequences of words. 
This idea can be promptly casted in our framework by predicting the next query in a session given the previous one.
With respect to our example, the query encoding estimated by the RNN in Figure~\ref{fig:rnn_intro}~(a) can be used as input to the RNN in Figure~\ref{fig:rnn_intro}~(b): the model learns a mapping between the consecutive queries \emph{cleveland gallery} and \emph{lake erie art}.  
At test time, the user query is encoded and then decoded into likely continuations that may be used as suggestions.

\begin{figure*}[thbp]
\centering
\includegraphics[scale=0.6]{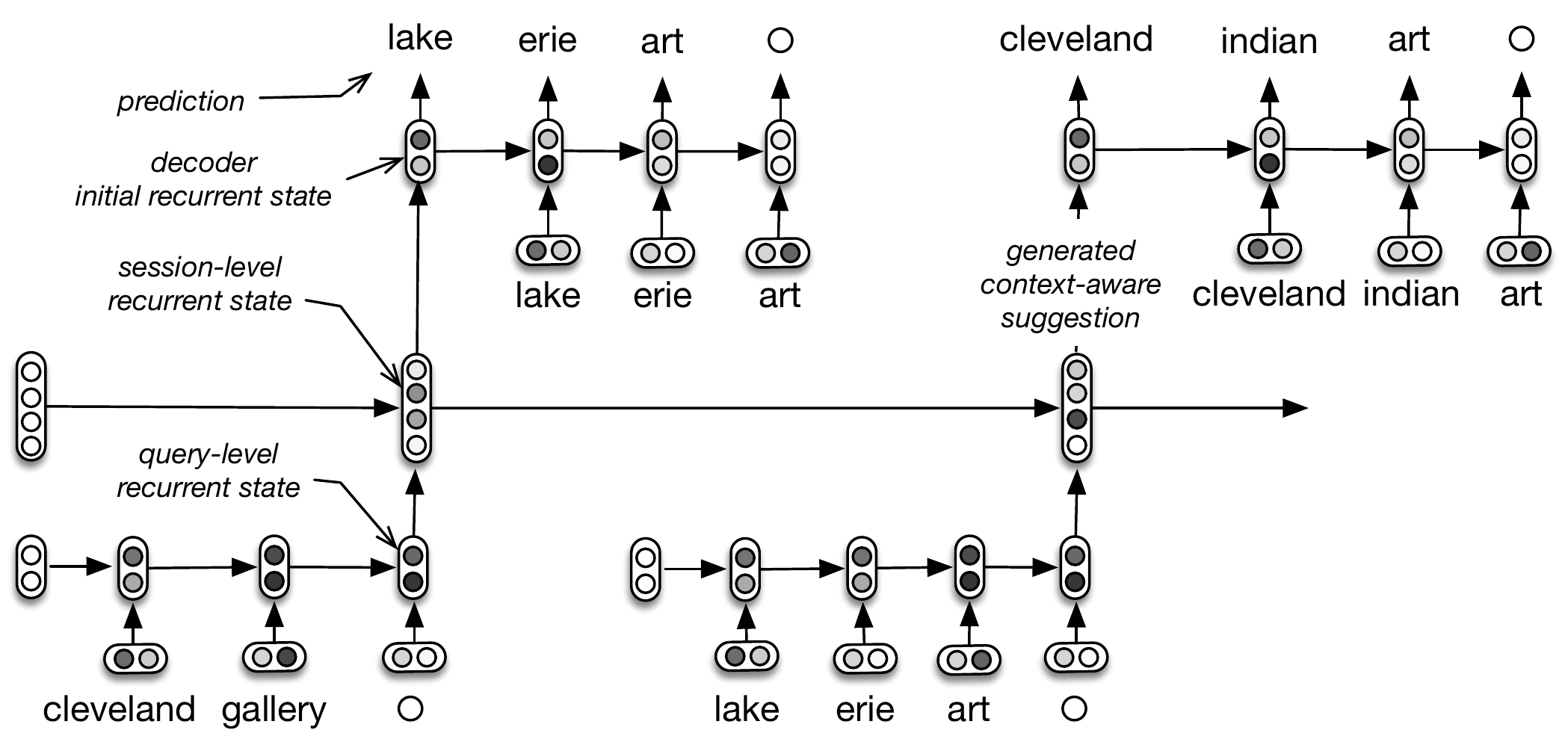}
\caption{\label{fig:hierarchical} The hierarchical recurrent encoder-decoder (HRED) for query suggestion. Each arrow is a non-linear transformation. The user types \emph{cleveland gallery} $\rightarrow$ \emph{lake erie art}. During training, the model encodes \emph{cleveland gallery}, updates the session-level recurrent state and maximize the probability of seeing the following query \emph{lake erie art}. The process is repeated for all queries in the session. During testing, a contextual suggestion is generated by encoding the previous queries, by updating the session-level recurrent states accordingly and by sampling a new query from the last obtained session-level recurrent state. In the example, the generated contextual suggestion is \emph{cleveland indian art}.}
\end{figure*}

Although powerful, such mapping is pairwise, and as a result, most of the query context is lost.
To condition the prediction of the next query on the previous queries in the session, we deploy an additional, session-level RNN on top of the query-level RNN encoder, thus forming a \emph{hierarchy} of RNNs (Figure~\ref{fig:hierarchical}).
The query-level RNN is responsible to encode a query.
The session-level RNN takes as input the query encoding and updates its own recurrent state.
At a given position in the session, the session-level recurrent state is a learnt summary of the past queries, keeping the information that is relevant to predict the next one.
At this point, the decoder RNN takes as input the session-level recurrent state, thus making the next query prediction contextual.

The contribution of this architecture is two-fold.
The query-level encoder RNN maps similar queries to vectors close in the embedding space (Figure~\ref{fig:word_query_embeddings}~(b)).
The mapping generalizes to queries that have not been seen in the training data, as long as their words appear in the model vocabulary.
This allows the model to map rare queries to more useful and general formulations, well beyond past co-occurred queries.
The session-level RNN models the sequence of the previous queries, thus making the prediction of the next query contextual.
Similar contexts are mapped close to each other in the vector space.
This property allows to avoid sparsity, and differently from count-based models~\cite{Cao2008,He2009}, to account for contexts of arbitrary length. 

\section{Mathematical Framework}
\label{sec:model}

We start by presenting the technical details of the RNN architecture, which our model extends. We consider a query session as a sequence of $M$ queries $S = \{ Q_1, \ldots, Q_M \}$ submitted by a user in chronological order, i.e. ${Q_m} <_t {Q_{m+1}}$ where $<_t$ is the total order generated by the submission time, and within a time frame, usually 30 minutes. A query $Q_m$ is a sequence of words $Q_m = \{ w_{m, 1}\,, \ldots\,, w_{m, N_m} \}$, where $N_m$ is the length of query $m$. $V$ is the size of the vocabulary.

\subsection{Recurrent Neural Network}
For each query word $w_n$, a RNN computes a dense vector called \emph{recurrent} state, denoted $h_n$, that combines $w_n$ with the information that has already been processed, i.e.~the recurrent state $h_{n - 1}$. Formally:
\begin{equation}
\label{eq:general_rec}
h_n = f(h_{n - 1}, w_n), \; h_0 = 0
\end{equation}
where $h_n \in \mathbb{R}^{d_h}$, $d_h$ is the number of dimensions of the recurrent state, $f$ is non-linear transformation and the recurrence is seeded with the 0 vector. The recurrent state $h_n$ acts as a compact \emph{summary} of the words seen up to position $n$. 

Usually, $f$ consists of a non-linear function, i.e.~the logistic sigmoid or hyperbolic tangent, 
applied element-wise to a time-independent affine transformation~\cite{Mikolov2010}.
The complexity of the function $f$ has an impact on how accurately the RNN can represent sentence information 
for the task at hand. To reduce the fundamental difficulty in learning long-term dependencies~\cite{bengio1994learning}, i.e.~to store information for longer sequences, more complex functions have been proposed such as the Long Short-Term Memory (LSTM)~\cite{hochreiter1997long} and the Gated Recurrent Unit (GRU)~\cite{Kyung14}.

Once Eq.~\ref{eq:general_rec} has been run through the entire query, the recurrent states $h_1, \ldots, h_N$ can be used in various ways. In an encoder RNN, the last state $h_N$ may be viewed as an order-sensitive compact summary of the input query. 
In a decoder RNN, the recurrent states are used to predict the next word in a sequence~\cite{Kyung14,Mikolov2010}. Specifically, the word at position $n$ is predicted using $h_{n - 1}$.
The probability of seeing word $v$ at position $n$ is:
\begin{equation}
\label{eq:shallow_output}
P(w_{n} = v | w_{1:n-1}) = \frac{\exp  o_{v}^\top h_{n - 1} }{\sum_{k} \exp  o_{k}^\top h_{n - 1} },
\end{equation}
where $o_i \in \mathbb{R}^{d_e}$ is a real-valued vector of dimensions $d_e$ associated to word $i$, i.e. a word embedding, and the denominator is a normalization factor. 
A representation of the embeddings learnt by our model is given in Figure~\ref{fig:word_query_embeddings}~(a).
The semantics of Eq.~\ref{eq:shallow_output} dictates that the probability of seeing word $v$ at position $n$ increases if its corresponding embedding vector $o_v$ is ``near'' the context encoded in the vector $h_{n - 1}$.
The parameters of the RNN are learned by maximizing the likelihood of the sequence, computed using Eq.~2.

\subsubsection{Gated Recurrent Unit}
\label{sec:gru}
We choose to use the Gated Recurrent Unit (GRU) as our non-linear transformation $f$. GRUs have demonstrated to achieve better performance than simpler parameterizations at an affordable computational cost~\cite{Kyung14}.
This function reduces the difficulties in learning our model by easing the propagation of the gradients.
We let $w_n$ denote the one-hot representation of $w_n = v$, i.e.~a vector of the size of the vocabulary with a 1 corresponding to the index of the query word $v$. The specific parameterization of $f$ is given by:
\begin{equation}
\begin{split}
\label{eq:gru}
& r_{n} = \sigma ( I_r w_n + H_r h_{n-1} ), \hspace{1.3cm} \emph{(reset gate)}\\
& u_{n} = \sigma ( I_u w_n + H_u h_{n-1} ), \hspace{1.2cm} \emph{(update gate)} \\
& \bar h_n = \text{tanh} ( I w_n + H (r_n \cdot h_{n-1}) ), \hspace{0.25cm} \emph{(candidate update)}\\ 
& h_n = (1-u_n) \cdot h_{n-1} + u_n \cdot \bar h, \hspace{0.7cm} \emph{(final update)}\\
\end{split}
\end{equation}
where $\sigma$ is the logistic sigmoid, $\sigma(x) \in [0, 1]$, $\cdot$ represents the element-wise scalar product between vectors, 
$I, I_u, I_r \in \mathbb{R}^{d_h \times V}$ and $H, H_r, H_u$ are in $\mathbb{R}^{d_h \times d_h}$. The $I$ matrices encode the word $w_n$ while the $H$ matrices specialize in retaining or forgetting the information in $h_{n-1}$. In the following, this function will be noted $GRU(h_{n-1}, w_n)$.

The gates $r_n$ and $u_n$ are computed in parallel.
If, given the current word, it is preferable to forget information about the past, i.e.~to reset parts of $h_n$, the elements of $r_n$ will be pushed towards 0.
The update gate $u_n$ plays the opposite role, i.e.~it judges whether the current word contains relevant information that should be stored in $h_n$.
In the final update, if the elements of $u_n$ are close to 0, the network discards the update $\bar h$ and keeps the last recurrent state $h_{n-1}$.
The gating behaviour provides robustness to noise in the input sequence: we hypothesize that this is particularly important for IR as it allows, for example, to exclude from the summary non-discriminative terms appearing in the query.

\subsection{Architecture}
Our hierarchical recurrent encoder-decoder (HRED) is pictured in Figure~\ref{fig:hierarchical}.
Given a query in the session, the model encodes the information seen up to that position and tries to predict the following query.
The process is iterated throughout all the queries in the session.
In the forward pass, the model computes the query-level encodings, the session-level recurrent states and 
the log-likelihood of each query in the session given the previous ones.
In the backward pass, the gradients are computed and the parameters are updated.
 
\subsubsection{Query-Level Encoding}
For each query $Q_m = \{w_{m,1}, \ldots, w_{m,N_m}\}$ in the training session $S$, the query-level RNN reads the words of the query sequentially and updates its hidden state according to:
\begin{equation}
\label{eq:encoder-step}
h_{m,n} = GRU_{enc}(h_{m,n - 1}, w_{m,n}), \;\; n = 1, \ldots, N_m,\;\;
\end{equation}
where $GRU_{enc}$ is the query-level encoder GRU function in Eq.~\ref{eq:gru}, $h_{m,n} \in \mathbb{R}^{d_h}$ and $h_{m,0} = 0$, the null vector. 
The recurrent state $h_{m, N_m}$ is a vector storing order-sensitive information about all the words in the query. 
To keep the notation uncluttered, we denote $q_m \equiv h_{m, N_m}$ the vector for query $m$. 
In summary, the query-level RNN encoder maps a query to a fixed-length vector.
Its parameters are shared across the queries.
Therefore, the obtained query representation $q_m$ is a general, 
acontextual representation of query $m$.
The computation of the $q_1, \ldots, q_M$ can be performed in parallel, thus lowering computational costs.
A projection of the generated query vectors is provided in Figure~\ref{fig:word_query_embeddings}~(b).

\subsubsection{Session-Level Encoding}
The session-level RNN takes as input the sequence of query representations $q_1, \ldots, q_M$ and computes the sequence of session-level recurrent states. For the session-level RNN, we also use the GRU function:
\begin{equation}
\label{eq:session-step}
s_m = GRU_{ses}(s_{m-1}, q_m), \;\; m = 1, \ldots, M,
\end{equation}
where $s_m \in \mathbb{R}^{d_s}$ is the session-level recurrent state, $d_s$ is its dimensionality and $s_0 = 0$. The number of session-level recurrent states $s_m$ is $M$, the number of queries in the session. 

The session-level recurrent state $s_m$ summarizes the queries that have been processed up to position $m$. 
Each $s_m$ bears a particularly powerful characteristic: it is sensitive to the order of previous queries and, as such, it can potentially encode order-dependent reformulation patterns such as generalization or specification of the previous queries~\cite{Huang2009}.
Additionally, it inherits from the query vectors $q_m$ the sensitivity to the order of words in the queries.

\subsubsection{Next-Query Decoding}
The RNN decoder is responsible to predict the next query $Q_{m}$ given the previous queries $Q_{1:m-1}$, i.e.~to estimate the probability:
\begin{equation}
\label{eq:prob_next_query}
P(Q_m | Q_{1:m-1}) = \prod_{n=1}^{N_m} P(w_n | w_{1:n-1}, Q_{1:m-1}).
\end{equation}
The desired conditioning on previous queries is obtained by initializing the recurrence of the RNN decoder with a non-linear transformation of $s_{m-1}$:
\begin{equation}
\label{eq:init-decoder}
d_{m, 0} = \text{tanh} ( D_0 s_{m-1} + b_0 ),
\end{equation}
where $d_{m, 0} \in \mathbb{R}^{d_h}$ is the decoder initial recurrent state (depicted in Figure~\ref{fig:hierarchical}), $D_0 \in \mathbb{R}^{d_h \times d_s}$ projects the context summary into the decoder space and $b_0 \in \mathbb{R}^{d_h}$.
This way, the information about previous queries is transferred to the decoder RNN.
The recurrence takes the usual form:
\begin{equation}
\label{eq:decoder-step}
d_{m, n} = GRU_{dec}(d_{m, n - 1} \,, w_{m, n}), \;\; n = 1, \ldots, N_{m},
\end{equation}
where $GRU_{dec}$ is the decoder GRU, $d_{m,n} \in \mathbb{R}^{d_h}$~\cite{Kyung14}.
In a RNN decoder, each recurrent state $d_{m, n-1}$ is used to compute the probability of the next word $w_{m, n}$.
The probability of word $w_{m, n}$ given previous words and queries is:
\begin{equation}
\label{eq:prob_word_next_query}
\begin{split}
P(w_{m, n} = v\, |\, & w_{m, 1:n-1},\, Q_{1:m-1}) = \\
& = \frac{\exp o_v^\top \omega ( d_{m, n-1}\,, w_{m, n-1} )}{\sum_{k} \exp o_k^\top \omega ( d_{m, n-1}\,, w_{m, n-1} )},
\end{split}
\end{equation}
where $o_v \in \mathbb{R}^{d_e}$ is the output embedding of word $v$ and $\omega$ is a function of both the recurrent state at position $n$ and the last input word: 
\begin{equation}
\omega ( d_{m, n - 1}, w_{m, n - 1} ) = H_o\, d_{m, n - 1} + E_o\, w_{m, n - 1} + b_o,
\end{equation}
where $H_o \in \mathbb{R}^{d_e \times d_h}$, $E_o \in \mathbb{R}^{d_e \times V}$ and $b_o \in \mathbb{R}^{d_e}$. To predict the first word of $Q_m$,
we set $w_{m, 0} = 0$, the 0 vector.
Instead of using the recurrent state directly as in Eq.~\ref{eq:shallow_output}, we add another layer of linear transformation $\omega$.
The $E_o$ parameter accentuates the responsibility of the previous word to predict the next one.
This formulation has shown to be beneficial for language modelling tasks~\cite{PascanuDeep13,Kyung14,Mikolov2010}.
If $o_v$ is ``near'' the vector $\omega ( d_{m, n-1}, w_{m, n-1} )$ the word $v$ has high probability under the model.

\subsection{Learning}
The model parameters comprise the parameters of the three GRU functions, $GRU_{enc}$, $GRU_{dec}$, $GRU_{ses}$, the output parameters $H_o, E_o, b_o$ and the $V$ output vectors $o_i$. 
These are learned by maximizing the log-likelihood of a session $S$, defined by the probabilities estimated with Eq.~\ref{eq:prob_next_query} and Eq~\ref{eq:prob_word_next_query}:
\begin{equation}
\begin{split}
\mathcal{L}(S) & =  \sum_{m = 1}^M \log P(Q_m | Q_{1:m-1}) \\ 
& = \sum_{m = 1}^M \sum_{n = 1}^{N_m} \log P(w_{m, n} | w_{m, 1:n-1}, Q_{1:m-1}).
\label{eq:likelihood}
\end{split}
\end{equation}
The gradients of the objective function are computed using the back-propagation through time (BPTT) algorithm~\cite{BPTT1988}.

\subsection{Generation and Rescoring}
\vspace{-3mm}
\label{sec:inference}
\paragraph*{Generation}
In our framework, the query suggestion task corresponds to an inference problem. A user submits the sequence of queries $S = \{Q_1, \ldots, Q_M\}$. A query suggestion is a query $Q^*$ such that:
\begin{equation}
\label{eq:inference}
Q^* = \arg\max_{Q \in \mathcal{Q}} P(Q |�Q_{1:M}),
\end{equation}
where $\mathcal{Q}$ is the space of possible queries, i.e.~
the space of sentences ending by the end-of-query symbol.
The solution to the problem can be approximated using standard word-level decoding techniques such as beam-search~\cite{Kyung14,Koehn2009}.
We iteratively consider a set of $k$ best prefixes up to length $n$ as candidates and
we extend each of them by sampling the most probable $k$ words given the distribution in Eq.~\ref{eq:prob_word_next_query}. We obtain $k^2$ queries of length $n+1$ and keep only the $k$ best of them. 
The process ends when we obtain $k$ well-formed queries containing the special end-of-query token $\circ$.

\begin{table}[t]
\renewcommand{\arraystretch}{1}
\centering
\small
\begin{tabular}[hbp]{p{4.2cm}p{3.5cm}}
\toprule
Context & Synthetic Suggestions \\ 
\midrule
ace series drive & ace hardware \\
& ace hard drive \\
& hp officejet drive \\
& ace hardware series \\
 \midrule
cleveland gallery $\rightarrow$ lake erie art & cleveland indian art \\
& lake erie art gallery \\
& lake erie picture gallery \\
& sandusky ohio art gallery \\
\bottomrule
\end{tabular}
\caption{\label{tab:samples}HRED suggestions given the context.}
\end{table}

\paragraph*{Example}
\vspace{-2mm}
Consider a user who submits the queries \emph{cleveland gallery} $\rightarrow$ \emph{lake erie artist}.
The suggestion system proceeds as follows.
We apply Eq.~\ref{eq:encoder-step} to each query obtaining the query vectors $q_{\text{cleveland gallery}}$ and $q_{\text{lake erie art}}$.
Then, we compute the session-level recurrent states by applying Eq.~\ref{eq:session-step} to the query vectors.
At this point, we obtain two session-level recurrent states, $s_{\text{cleveland gallery}}$ and $s_{\text{lake erie art}}$.
To generate context-aware suggestions, we start by mapping the last session-level recurrent state, $s_{\text{lake erie art}}$, into the initial decoder input $d_0$ using Eq.~\ref{eq:init-decoder}.
We are ready to start the sampling of the suggestion. 
Let assume that the beam-search size is $1$.
The probability of the first word $w_1$ in the suggestion is computed using Eq.~\ref{eq:prob_word_next_query} by using $d_0$ and $w_0 = 0$, the null vector.
The word with the highest probability, i.e.~\emph{cleveland}, is added to the beam. 
The next decoder recurrent state $d_1$ is computed by means of Eq.~\ref{eq:decoder-step} using $d_0$ and $w_{1} = \emph{cleveland}$.
Using $d_1$, we are able to pick $w_2 = \emph{indian}$ as the second most likely word.
The process repeats and the model selects \emph{art} and $\circ$.
As soon as the end-of-query symbol is sampled, the context-aware suggestion \emph{cleveland indian art} is presented to the user. In Table~\ref{tab:samples} we give an idea of the generated suggestions for 2 contexts in our test set.

\paragraph*{Rescoring}
\vspace{-2mm}
\label{sec:scoring}
Our model can evaluate the likelihood of a given suggestion conditioned on the history of previous queries through Eq.~\ref{eq:prob_next_query}.
This makes our model integrable into more complex suggestion systems.
In the next section, we choose to evaluate our model by adding the likelihood scores of candidate suggestions as additional features into a learning-to-rank system.
\vspace{-2mm}

\section{Experiments}
\label{sec:experiment}

We test how well our query suggestion model can predict the next query in the session given the history of previous queries.
This evaluation scenario aims at measuring the ability of a model to propose the target next query, which is assumed to be one desired by the user.
We evaluate this with a learning-to-rank approach (explained in Section~\ref{sec:ltr}), similar to the one used in~\cite{Bhaskar15,Shok2013} for query auto-completion and in~\cite{Ozertem2012,Santos2013} for query suggestion.
We first generate candidates using a co-occurrence based suggestion model.
Then, we train a baseline ranker comprising a set of contextual features depending on the history of previous queries as well as pairwise features which depend only on the most recent query.
The likelihood scores given by our model are used as additional features in the supervised ranker.
At the end, we have three systems:
(1) the original co-occurrence based ranking, denoted ADJ;
(2) the supervised context-aware ranker, which we refer to as Baseline Ranker;
and (3) a supervised ranker with our HRED feature.
We evaluate the performance of the model and the baselines using mean reciprocal rank (MRR).
This is common for tasks whose ground truth is only one instance~\cite{Jiang2014,Bhaskar15}.

\subsection{Dataset}
We conduct our experiments on the well-known search log from AOL, which is the only 
available search log that is large enough to train our model and the baselines. 
The queries in this dataset were sampled between 1 March, 2006 and 31 May, 2006. 
In total there are 16,946,938 queries submitted by 657,426 unique users.
We remove all non-alphanumeric characters from the queries, apply a spelling corrector and lowercasing.
After filtering, we sort the query log by query timestamp and we use the queries submitted before 1 May, 2006 as our \emph{background} data to estimate the proposed model and the baselines.
The next two weeks of data are used as a \emph{training} set for tuning the ranking models.
The remaining two weeks are split into the \emph{validation} and the \emph{test} set.
We follow common practice and we define the end of a session by a 30 minute window of idle time~\cite{Jansen2007}.
After filtering, there are 1,708,224 sessions in the background set, 435,705 in the training set, 166,836 in the validation set and 230,359 sessions in the testing set. 

\begin{table}
\centering
\small
\begin{tabular}[h]{cccc}
\toprule
Batches Seen & Training & Decoding ($50$) & Memory \\ 
\midrule
135,350 & 44h\,01m & $\sim$ 1s & 301\,Mb \\
\bottomrule
\end{tabular}
\caption{\label{tab:training}Full statistics about training time, memory impact and decoding time with a beam size of 50.}
\end{table}

\subsection{Model Training}
The most frequent $90K$ words in the background set form our vocabulary $V$.
This is a common setting for RNN applied to language and allows to speed-up the repeated summations over $V$ in Eq.~\ref{eq:prob_word_next_query}~\cite{Kyung14,Ilya14}.
Parameter optimization is done using mini-batch RMSPROP~\cite{Graves13}.
We stabilize the learning by normalizing the gradients if their norm exceeds a threshold $c = 1$~\cite{PascanuDeep13}.
The training stops if the likelihood of the validation set does not improve for $5$ consecutive iterations.
We train our model using the Theano\footnote{An implementation of the model is available at \texttt{https://github.com/sordonia/hed-qs}.} library~\cite{Theano12,Theano10}.
The dimensionality of the query-level RNN is set to $d_h = 1000$.
To ensure a high-capacity session-level RNN, we set $d_s = 1500$.
This is useful to memorize complex information about previous queries.
The output word embeddings $o_i$ are 300 dimensional vectors, i.e.~$d_e = 300$.
Differently from context-aware approaches for which the model size increases with the number of queries, our model is compact and can easily fit in memory (Table~\ref{tab:training}). 

\subsection{Learning to Rank}
\label{sec:ltr}
Given a session $S = \{Q_1, \ldots, Q_M\}$, we aim to predict the \emph{target} query $Q_M$ given the \emph{context} $Q_1, \ldots, Q_{M-1}$. $Q_{M-1}$ is called the \emph{anchor} query and will play a crucial role in the selection of the candidates to rerank.
To probe different capabilities of our model, we predict the next query in three scenarios:
(a) when the anchor query exists in the background data (Section~\ref{sec:general_setting});
(b) when the context is perturbed with overly common queries (Section~\ref{sec:robust_setting}); 
(c) when the anchor is not present in the background data (Section~\ref{sec:longtail_setting}).

For each session, we select a list of $20$ possible candidates to rerank.
The exact method used to produce the candidates will be discussed in the next sections. 
Once the candidates are extracted, we label the true target as relevant and all the others as non-relevant.
We choose to use one of the state-of-the-art  ranking algorithms LambdaMART as our supervised ranker, which is the winner in the Yahoo! Learning to Rank Challenge in 2010~\cite{mart}. We tune the LambdaMART model with 500 trees and the parameters are learnt using standard separate training and validation set.


We describe the set of pairwise and contextual features (17 in total) used to train a supervised baseline prediction model, denoted Baseline Ranker. The baseline ranker is a competitive system comprising features that are comparable with the ones described in the literature for query auto-completion~\cite{Jiang2014,Bhaskar15} and next-query prediction~\cite{He2009}.
\vspace{-2mm}
\paragraph*{Pairwise and Suggestion Features} For each candidate suggestion, we count how many times it follows the anchor query in the background data and add this count as a feature. Additionally, we use the frequency of the anchor query in the background data. Following~\cite{Jiang2014,Ozertem2012} we also add the Levenshtein distance between the anchor and the suggestion. Suggestion features include: the suggestion length (characters and words) and its frequency in the background set. 

\vspace{-2mm}
\paragraph*{Contextual Features} 
Similarly to~\cite{Bhaskar15,Shok2013}, we add 10 features corresponding to the character $n$-gram similarity between the suggestion and the 10 most recent queries in the context.
We add the average Levenshtein distance between the suggestion and each query in the context~\cite{Jiang2014}. 
We use the scores estimated using the context-aware Query Variable Markov Model (QVMM)~\cite{He2009} as an additional feature.
QVMM models the context with a variable memory Markov model able to automatically back-off shorter query $n$-grams if the exact context is not found in the background data.

\begin{figure}
\centering
\includegraphics[scale=0.50]{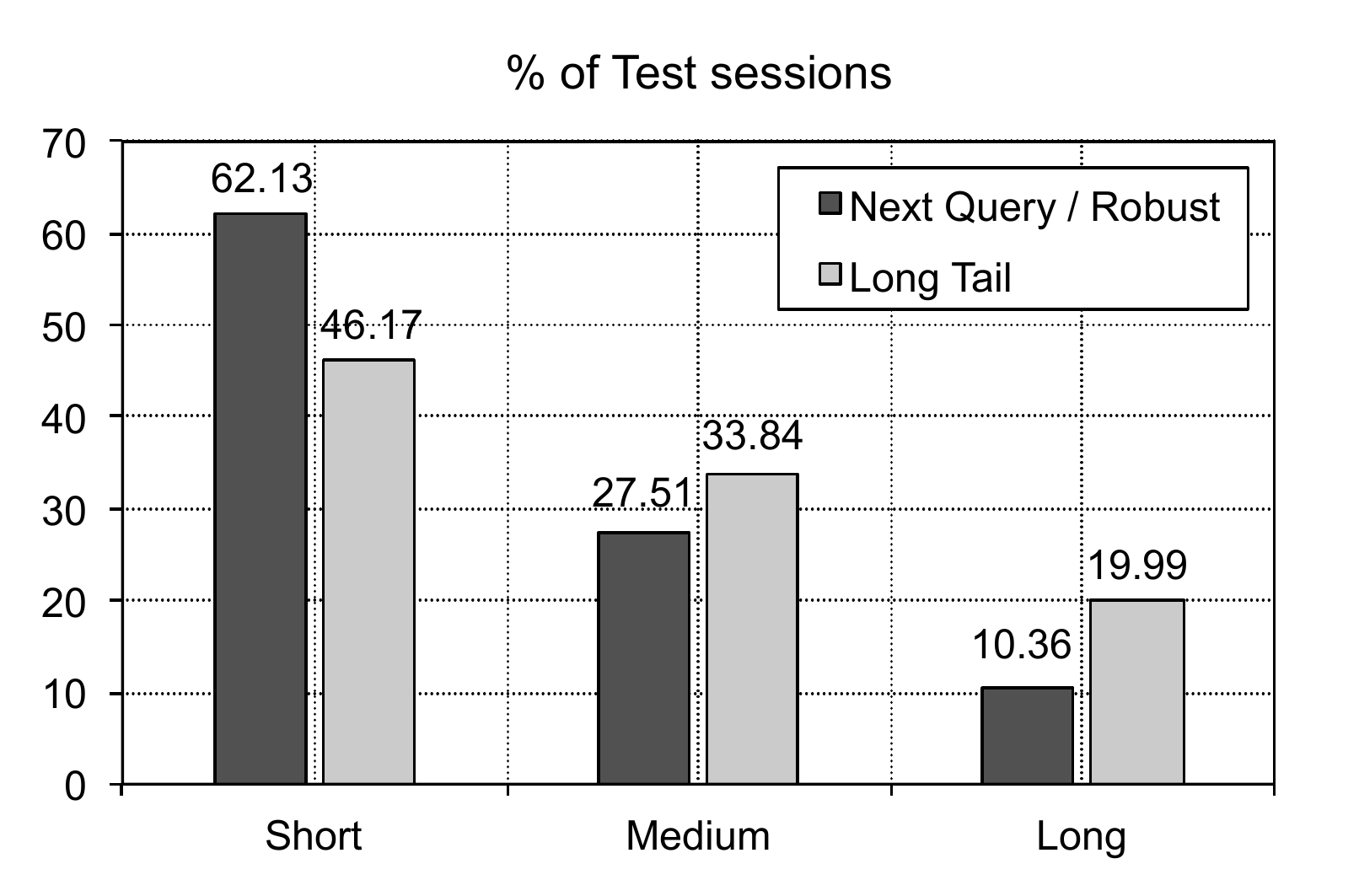}
\vspace{-2mm}
\caption{\label{fig:distribution_length} Proportion (\%) of short (2 queries), medium (3 or 4 queries) and long (at least 5 queries) sessions in our test scenarios.}
\end{figure}

\vspace{-2mm}
\paragraph*{HRED Score}
The proposed Hierarchical Recurrent Encoder Decoder (HRED) contributes one additional feature corresponding to the log-likelihood of the suggestion given the context, as detailed in Section~\ref{sec:scoring}.

\subsection{Test Scenario 1: Next-Query Prediction}
\label{sec:general_setting}
For each session in the training, validation and test set, we extract 20 queries that most likely follow the anchor query in the background data, i.e.~with the highest ADJ score.
The session is included if and only if at least 20 queries have been extracted and the target query appears in the candidate list.
In that case, the target query is the positive candidate and the 19 other candidates are the negative examples.
Note that a similar setting has been used in~\cite{Jiang2014,Bhaskar15} for query auto-completion.
We have 18,882 sessions in the training, 6,988 sessions in the validation and 9,348 sessions in the test set.
The distribution of the session length is reported in Figure~\ref{fig:distribution_length}.
The scores obtained by the ADJ counts are used as an additional non-supervised baseline.

\begin{table}[t]
\centering
\renewcommand{\arraystretch}{1.1}
\begin{tabular}{lcccl}
\toprule
Method & \hspace{1.5cm} & MRR & & $\Delta$\% \\
\midrule
 ADJ & & 0.5334  & & - \\
Baseline Ranker & & $0.5563$  & & \emph{+4.3\%} \\
+ HRED            & & \textbf{0.5749} & & \emph{+7.8\%/+3.3\%}  \\
\bottomrule
\end{tabular}
\caption{\label{tab:next_query}Next-query prediction results. All improvements are significant by the t-test ($p$ < 0.01).}
\end{table}

\paragraph*{Main Result}

Table~\ref{tab:next_query} shows the MRR performance for our model and the baselines.
Baseline Ranker achieves a relative improvement of $4.3\%$ with respect to the ADJ model. 
We find that the HRED feature brings additional gains achieving $7.8\%$ relative improvement over ADJ.
The differences in performance with respect to ADJ and the Baseline Ranker are significant using a t-test with $p < 0.01$.
In this general next-query prediction setting, HRED boosts the rank of the first relevant result.

\begin{figure}[t]
\centering
\includegraphics[scale=0.38]{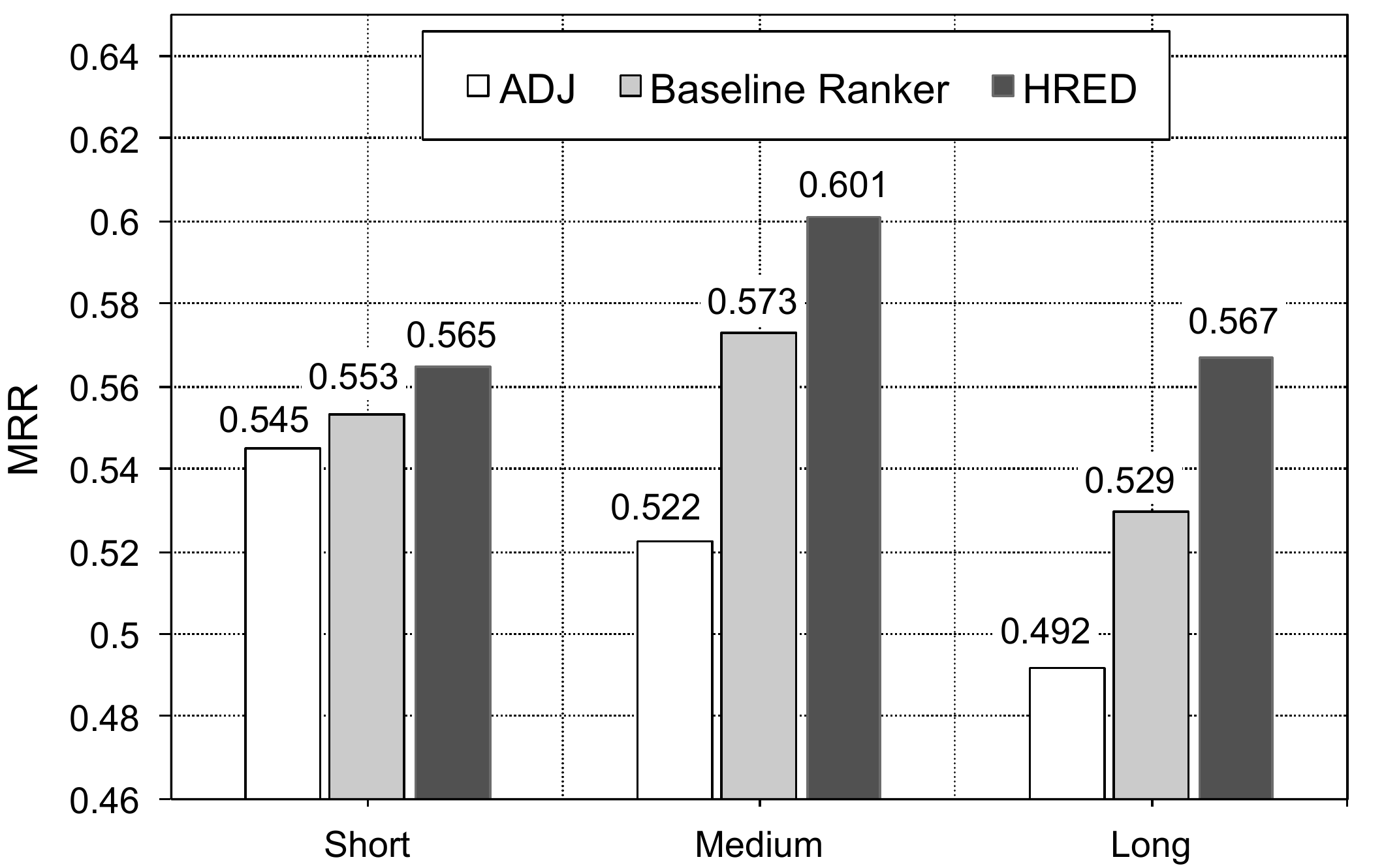}
\caption{\label{fig:next_query_by_length}  Next-query performance in short, medium and long sessions. All differences in MRR are statistically significant by the t-test ($p$ < 0.01).}
\end{figure}

\begin{figure}[t]
\centering
\includegraphics[scale=0.42]{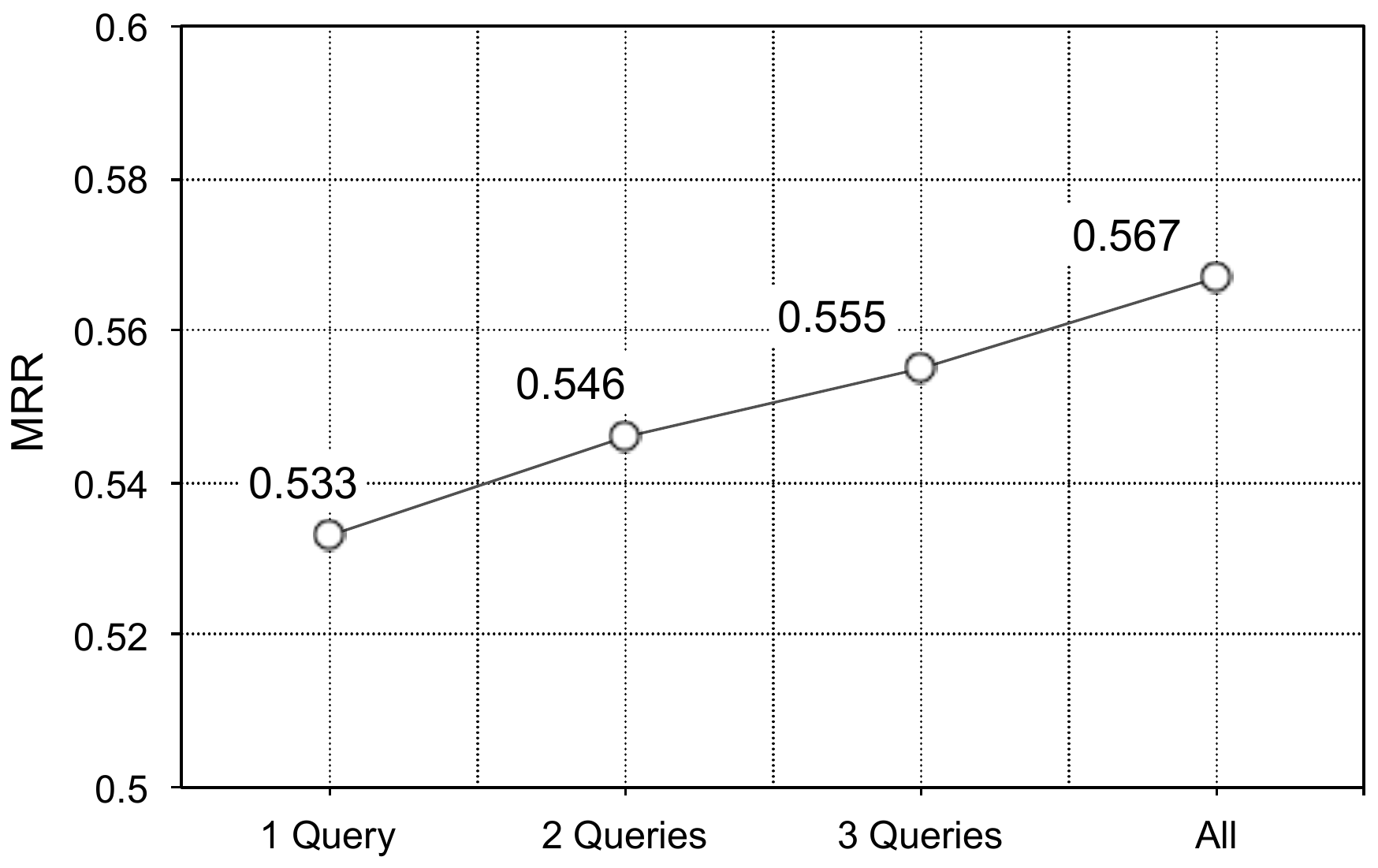}
\vspace{-2mm}
\caption{\label{fig:long_by_context_length} Variation of HRED performance with respect to the number of previous queries considered. The evolution is computed on long sessions.}
\end{figure}

\vspace{-2mm}
\paragraph*{Impact of Session Length} We expect the session length to have an impact on the performance of context-aware models.
In Figure~\ref{fig:next_query_by_length}, we report separate results for short (2 queries), medium (3 or 4 queries) and long sessions (at least 5 queries).
HRED brings statistically significant improvements across all the session lengths. 
For short sessions, the improvement is marginal but consistent even though only a short context is available in this case.
The semantic mapping learnt by the model appears to be useful, even in the pairwise case.
ADJ is affected by the lack of context-awareness and suffers a dramatic loss of performance with increasing session length.
In the medium range, context-aware models account for previous queries and achieve the highest performance.
The trend is not maintained for long sessions, seemingly the hardest for the Baseline Ranker.
Long sessions can be the result of complex search tasks involving a topically broad information need or changes of search topics.
Beyond the intrinsic difficulty in predicting the target query in these cases, exact context matches may be too coarse to infer the user need.
Count-based methods such as QVMM meet their limitations due to data sparsity.
In this difficult range, HRED achieves its highest relative improvement with respect to both ADJ (+15\%) and the Baseline Ranker (+7\%), thus showing robustness across different session lengths.
\vspace{-2mm}
\paragraph*{Impact of Context Length} We test whether the performance obtained by HRED on long sessions can be obtained using a shorter context. 
For each long session in our test set, we artificially truncate the context to make the prediction depend on the anchor query, $Q_{M-1}$, only (1 query), on $Q_{M-2}$ and $Q_{M-1}$ (2 queries), on 3 queries and on the entire context.
When one query is considered, our model behaves similarly to a pairwise recurrent encoder-decoder model trained on consecutive queries.
Figure~\ref{fig:long_by_context_length} shows that when only one query is considered, the performance of HRED is similar to the Baseline Ranker (0.529) which uses the whole context.
However, HRED appears to perform best when the whole context is considered, which highlights the importance of context-information.
Additional gains can be obtained by considering more than 3 queries, which highlights the ability of our model to consider long contexts.

\subsection{Test Scenario 2: Robust Prediction}
\label{sec:robust_setting}
Query sessions contain a lot of common and navigational queries such as \emph{google} or \emph{facebook} which do not correspond to a specific search topic. 
A context-aware suggestion system should be \emph{robust} to noisy queries and learn to discard them from the relevant history that should be retained. 
We propose to probe this capability by formulating an original robust prediction task as follows.
We label the 100 most frequent queries in the background set as \emph{noisy}\footnote{A similar categorization has been proposed in~\cite{Raman2014}.}.
For each entry in the training, validation and test set of the previous next-query prediction task, we corrupt its context by \emph{inserting} a noisy query at a random position.
The candidates and the target rest unchanged.
The probability of sampling a noisy query is proportional to its frequency in the background set.
For example, given the context \emph{airlines} $\rightarrow$ \emph{united airlines} and the true target \emph{delta airlines}, the noisy sample \emph{google} is inserted at a random position, forcing the models to predict the target given the corrupted context \emph{airlines} $\rightarrow$ \emph{united airlines} $\rightarrow$ \emph{google}.

\begin{table}[t]
\centering
\renewcommand{\arraystretch}{1.1}
\begin{tabular}{lcccl}
\toprule
Method & \hspace{1.5cm} & MRR & & $\Delta\%$ \\
\midrule
 ADJ & & 0.4507  & & - \\
Baseline Ranker & & $0.4831$  & & \emph{+7,2\%} \\
+ HRED            & & \textbf{0.5309} & & \emph{+17,8\%/+9.9\%}\\
\bottomrule
\end{tabular}
\caption{\label{tab:robust}Robust prediction results. The improvements are significant by the t-test ($p$ < 0.01).}
\end{table}

\begin{figure}[t]
\centering
\includegraphics[scale=0.45]{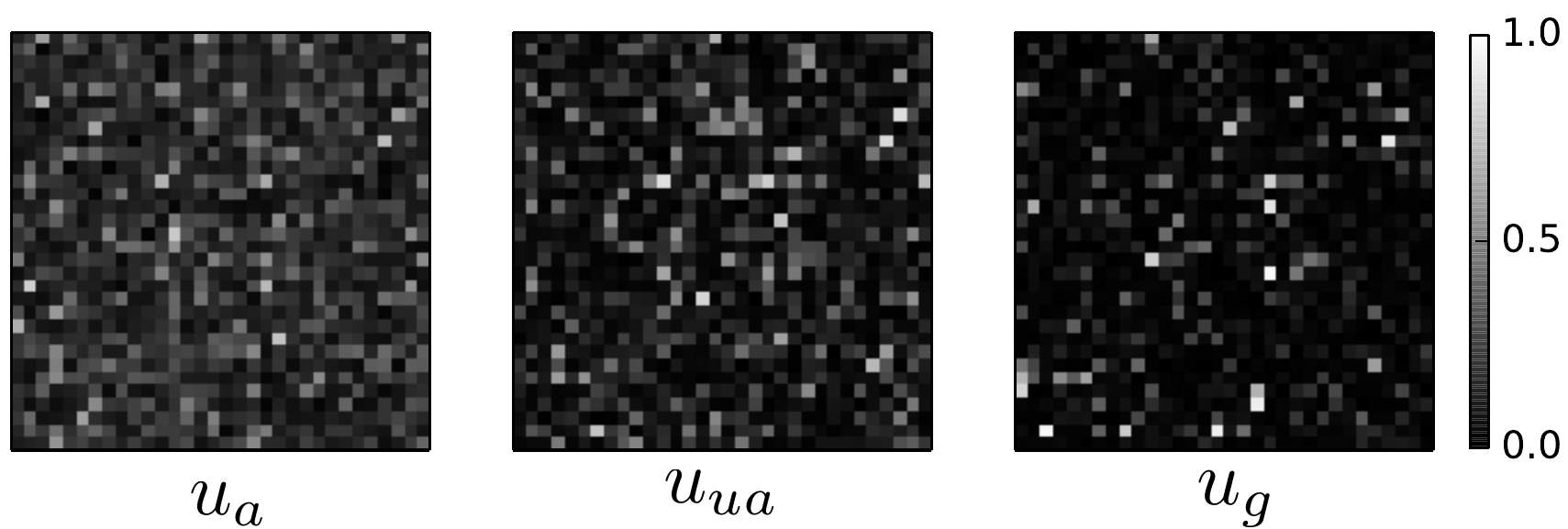}
\vspace{-2mm}
\caption{\label{fig:update_gates}Magnitude of the elements in the session-level update gates. The darker the image, the more the model discards the current query.
The vector corresponding to \emph{google}, $u_{g}$, is darker, i.e.~the network mainly keeps its previous recurrent state.}
\end{figure}

\vspace{-2mm}
\paragraph*{Main Result} Table~\ref{tab:robust} shows that corruption considerably affects the performance of ADJ.
Cases in which the corruption occurred at the position of the anchor query severely harm pairwise models.
The Baseline Ranker achieves significant gains over ADJ by leveraging context matches.
Its performance is inferior to the baseline ADJ performance in the next-query setting reported in Table~\ref{tab:next_query} (0.5334).
HRED appears to be particularly effective in this difficult setting achieving a relative improvement of $17.8\%$ over ADJ and $9.9\%$ over the Baseline Ranker, both statistically significant.
Comparative to the next-query task, the improvements over ADJ and the Baseline Ranker are 2.5 and 3 times higher respectively.
Our model appears to be more robust than the baselines in these extreme cases and can better reduce the impact of the noisy query.
\vspace{-2mm}
\paragraph*{Impact of the Hierarchical Structure}
As noisy queries bring little information to predict future queries in the session, HRED may automatically learn to be robust to the noise at training time.
The hierarchical structure allows to decide, for each query, if it is profitable to account for its contribution to predict future queries.
This capability is sustained by the session-level GRU, which can ignore the noisy queries by ``turning-off'' the update gate $u_n$ when they appear (see Section~\ref{sec:gru}).
Given the corrupted context \emph{airlines} $\rightarrow$ \emph{united airlines} $\rightarrow$ \emph{google}, the session-level GRU computes three update gate vectors: $u_{a}$, $u_{ua}$, $u_{g}$, each corresponding to a position in the context. 
In Figure~\ref{fig:update_gates}, we plot the magnitude of the elements in these vectors.
As the model needs to memorize the initial information, $u_{a}$ shows a significant number of non-zero (bright) entries.
At this point, general topical information has already been stored in the first recurrent state.
Hence, $u_{ua}$ shows a larger number of zero (dark) entries.
When \emph{google} is processed, the network tends to keep past information in memory by further zeroing entries in the update gate.
This sheds an interesting perspective: this mechanism may be used to address other search log related tasks such as session-boundary detection.

\subsection{Test Scenario 3: Long-Tail Prediction}
\label{sec:longtail_setting}
To analyze the performance of the models in the long-tail, we build our training, validation and test set by retaining the sessions for which the anchor query has not been seen in the background set, i.e.~it is a long-tail query.
In this case, we cannot leverage the ADJ score to select candidates to rerank.
For each session, we iteratively shorten the anchor query by dropping terms until we have a query that appears in the background data.
If a match is found, we proceed as described in the next-query prediction setting, that is, we guarantee that the target appears in the top 20 queries that have the highest ADJ scores given the anchor prefix.
The statistics of the obtained dataset are reported in Figure~\ref{fig:distribution_length}.
As expected, the distribution of lengths changes substantially with respect to the previous settings.
Long-tail queries are likely to appear in medium and long sessions, in which the user strives to find an adequate textual query.
\vspace{-2mm}
\paragraph*{Main Result} 
Table~\ref{tab:long_tail} shows that, due to the anchor prefix matching, ADJ suffer a significant loss of performance.
The performances of the models generally confirm our previous findings. 
HRED improves significantly by 5.6\% over the Baseline Ranker and proves to be useful even for long-tail queries.
Supervised models appear to achieve higher absolute scores in the long-tail setting than in the general next-query setting reported in Table~\ref{tab:next_query}.
After analysis of the long-tail testing set, we found that only~8\% of the session contexts contain at least one noisy query.
In the general next-query prediction case, this number grows to~37\%.
Noisy queries generally harm performance of the models by increasing the ambiguity in the next query prediction task.
This fact may explain why the Baseline ranker and HRED perform better on long-tail queries than in the general case.
It is interesting to see how the improvement of HRED with respect to the Baseline Ranker is larger for long-tail queries than in the general setup (5.6\% to 3.3\%).
Although not explicitly reported, we analyzed the performance with respect to the session length in the long-tail setting.
Similarly to the general next-query prediction setting, we found that the Baseline Ranker suffers significant losses for long sessions while our model appears robust to different session lengths.

\begin{table}
\centering
\renewcommand{\arraystretch}{1.1}
\begin{tabular}{lcccl}
\toprule
Method & \hspace{1.3cm} & MRR & & $\Delta$\% \\
\midrule
 ADJ & & 0.3830  & & - \\
Baseline Ranker & & $0.6788$  & & \emph{+77.2\%} \\
+ HRED            & & \textbf{0.7112} & & \emph{+85.3\%}/\emph{+5.6\%}\\
\bottomrule
\end{tabular}
\caption{\label{tab:long_tail}Long-tail prediction results. The improvements are significant by the t-test ($p$ < 0.01).}
\end{table}

\subsection{User Study}
\begin{figure}
\centering
\includegraphics[scale=0.45]{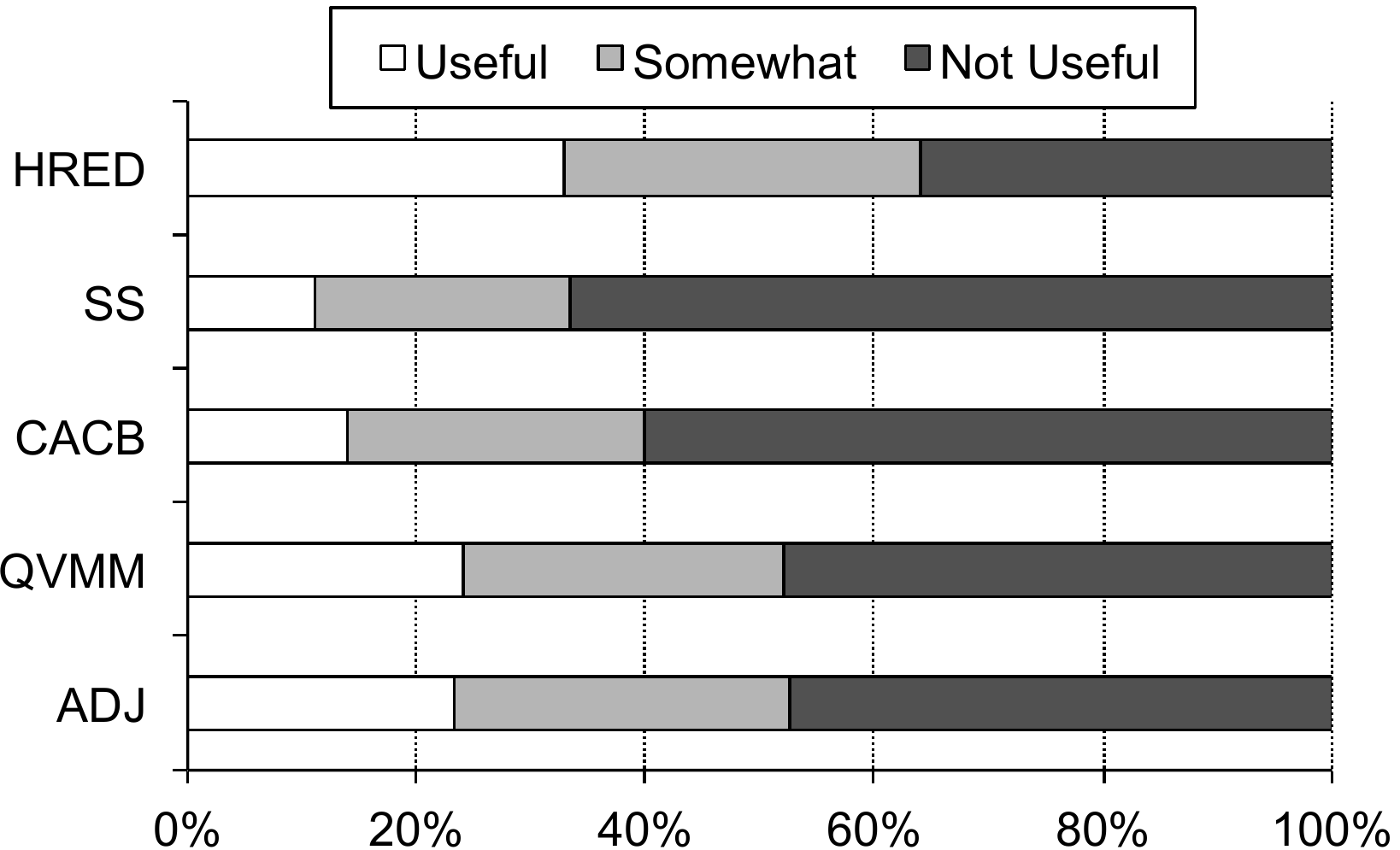}
\caption{\label{tab:user-study}User study results, which compare the
effectiveness of HRED with the baseline techniques.}
\end{figure}

The previous re-ranking setting doesn't allow to test the generative capabilities of our suggestion system.
We perform an additional user study and ask human evaluators to assess the quality of synthetic suggestions.
To avoid sampling bias towards overly common queries, we choose to generate suggestions for the 50 topics of the TREC Web Track 2011~\cite{clarke2011overview}.
The assessment was conducted by a group of 5 assessors.
To palliate the lack of context information for TREC queries, we proceed as follows: for each TREC topic $Q_M$, we extract from the test set the sessions ending exactly with $Q_M$ and we take their context $Q_1, \ldots, Q_{M-1}$. After contextualization, 19 TREC queries have one or more queries as context and the remaining are singletons.
For HRED, we build synthetic queries following the generative procedure described in Section~\ref{sec:inference}.
In addition to QVMM and ADJ, we compare our model with two other baselines: CACB~\cite{Cao2008}, which is similar to QVMM but builds clusters of queries to avoid sparsity, and SS (Search Shortcuts)~\cite{Broccolo2012}, which builds an index of the query sessions and extracts the last query of the most similar sessions to the source context.
Note that we do not compare the output of the previous supervised rankers as this would not test the generative capability of our model.
Each assessor was provided with a random query from the test bed, its context, if any, and a list of recommended queries (the top-5 for each of the methods) selected by the different methods. 
Recommendations were randomly shuffled, so that the assessors could not distinguish which method produced them. 
Each assessor was asked to judge each recommended query using the following scale: useful, somewhat useful, and not useful.
The user study finished when each assessor had assessed all recommendations for all 50 queries in the test bed. Figure~\ref{tab:user-study} reports the results of the user study averaged over all raters. Overall, for HRED, 64\% of the recommendations were judged useful or somewhat useful. The quality of the queries recommended by HRED is higher than our baselines both in the somewhat and in the useful category. 

\section{Related Works}
\label{sec:relatedwork}
\vspace{-3mm}
\paragraph*{Query Suggestion}
A notorious context-aware method was proposed by
He et al.~\cite{He2009}. The authors use a Variable Memory Markov model (QVMM) and build a suffix tree to model the user query sequence.
We used this model as a context-aware baseline feature in our supervised ranker.
The method by Cao et al.~\cite{Cao2008} is similar but they build a suffix tree on clusters of queries and model the transitions between clusters.
We didn't notice any improvements by adding this model as a feature in our case.
For both models, the number of parameters increases with the depth of the tree inducing sparsity.
Instead, our model can consider arbitrary length contexts with a fixed number of parameters.
Jiang et al.~\cite{Jiang2014} and Shokouhi et al.~\cite{Shok2013} propose context-aware approaches for query auto-completion.
We adopted a similar framework for query suggestion and use our model as a feature to rank the next-query.
Santos et al.~\cite{Santos2013} and Ozertem et al.~\cite{Ozertem2012} also use learning to rank approach for query suggestion.
In those cases, the rankers are trained using pairwise features and do not consider previous queries.
Interestingly, the authors model explicitly the usefulness of a suggestion by using click data and the result list.
In the future, we plan to integrate click information in the generation process of our model.

Query suggestion algorithms use clustering methods to find similar queries so that they can be used as suggestions for one another~\cite{BaezaYates2004, Wen2001}.
We demonstrated that our model exhibits similar clustering properties due to the embeddings learnt by the neural network.
Other works build a Query Flow Graph (QFG) to capture high-order query co-occurrence~\cite{Boldi2008, Sadikov2010}. 
Operating at the query-level, these methods suffer from the long-tail problem.
Bonchi et al.~\cite{Bonchi2012} propose a solution to these problems by introducing the Term-QFG (TQG), where single query terms are also included into the graph.
However, suggestion requires repeated complex random walks with restart. 
Similarly, our model can handle rare queries as long as their words appear in the model vocabulary. 
Vahabi et al.~\cite{Vahabi2013} find suggestions to long-tail queries by comparing their search results. 
Although effective, the approach requires to have $100$ results per query. 
A related approach is the Search Shortcut~ \cite{Broccolo2012} which avoids the long-tail problem by means of a retrieval algorithm.

Few synthetic suggestion models have been proposed in the literature. Szpektor et al.~\cite{Templates11} use a template generation
method by leveraging WordNet. Jain et al.~\cite{Jain2011} combine different resources and use a machine learning approach to prune redundant suggestions. These methods achieve automatic addition, removal and substitution of related terms into the queries. By maximizing the likelihood of the session data, our model learns to  perform similar modifications.

\vspace{-2mm}
\paragraph*{Neural Networks for NLP}
Neural networks have found several applications in a variety of tasks, ranging from Information Retrieval (IR)~\cite{huang2013learning,shen2014latent}, Language Modeling (LM)~\cite{Mikolov2010,Pascanu13} and Machine Translation (MT)~\cite{Kyung14,Ilya14}.
Cho et al.~\cite{Kyung14} and Sutskever et al.~\cite{Ilya14} use a Recurrent Neural Network (RNN) for end-to-end MT.
Our model bears similarities to these approaches but we contribute with the hierarchical structure.
The idea of encoding hierarchical multi-scale representations is also explored in \cite{El1995hierarchical}.
In IR, neural networks embeddings were used by Li et al~\cite{lideep}. The authors used deep feed-forward neural networks to use previous queries by the same user to boost document ranking. In~\cite{huang2013learning, shen2014latent}, the authors propose to use clickthrough data to learn a ranking model for ad-hoc IR.
Our model shares similarities with the interesting recent work by Mitra~\cite{Bhaskar15}.
The authors apply the discriminative pairwise neural model described in~\cite{shen2014latent} to measure similarity between queries.
Context-awareness is achieved at ranking time, by measuring the similarity between the candidates and each query in the context.
Our work has several key differences. First, we deploy a novel RNN architecture. Second, our model is generative. Third, we model the session context at training time.
To our knowledge, this is the first work applying RNNs to an IR task.

\section{Conclusion}
In this paper, we formulated a novel hierarchical neural network architecture and used it to produce query suggestions.
Our model is context-aware and it can handle rare queries.
It can be trained end-to-end on query sessions by simple optimization procedures.
Our experiments show that the scores provided by our model help improving MRR for next-query ranking. 
Additionally, it is generative by definition.
We showed with a user study that the synthetic generated queries are better than the compared methods. 

In future works, we aim to explicitly capture the usefulness of a suggestion by exploiting user clicks~\cite{Ozertem2012}.
This may be done without much effort as our architecture is flexible enough to allow joint training of other differentiable loss functions.
Then, we plan to further study the synthetic generation by means of a large-scale automatic evaluation.
Currently, the synthetic suggestions tend to be \emph{horizontal}, i.e.~the model prefers to add or remove terms from the context queries and rarely proposes orthogonal but related reformulations~\cite{Jain2011,Vahabi2013}.
Future efforts may be dedicated to diversify the generated suggestions to account for this effect.
Finally, the interactions of the user with previous suggestions can also be leveraged to better capture the behaviour of the user and to make better suggestions accordingly. 
We are the most excited about possible future applications beyond query suggestion: auto-completion, next-word prediction and other NLP tasks such as Language Modelling may be fit as possible candidates.
\vspace{-2mm}
\section*{Acknowledgments}
We would like to thank Jianfeng Gao, \c{C}a\u{g}lar G\"ul\c{c}ehre and Bhaskar Mitra for their precious advice, enlightening discussions and invaluable moral support. We gratefully acknowledge the support of NVIDIA Corporation with the donation of the Tesla K40 GPU used for this research.

\bibliographystyle{plain}
{\scriptsize
\vspace{-2mm}
\bibliography{cikm2015-sordonia}}

\begin{thebibliography}{10}

\bibitem{BaezaYates2004}
R.~Baeza-Yates, C.~Hurtado, and M.~Mendoza.
\newblock Query recommendation using query logs in search engines.
\newblock In {\em In Proc. of Int. Conf. on Current Trends in Database Tech.},
  pages 588--596, 2004.

\bibitem{Theano12}
F.~Bastien, P.~Lamblin, R.~Pascanu, J.~Bergstra, I.~J. Goodfellow, A.~Bergeron,
  N.~Bouchard, and Y.~Bengio.
\newblock Theano: new features and speed improvements.
\newblock Deep Learning and Unsupervised Feature Learning NIPS 2012 Workshop,
  2012.

\bibitem{Bengio2013}
Y.~Bengio.
\newblock Deep learning of representations: Looking forward.
\newblock {\em CoRR}, abs/1305.0445, 2013.

\bibitem{bengio2003neural}
Y.~Bengio, R.~Ducharme, and P.~Vincent.
\newblock A neural probabilistic language model.
\newblock {\em Journal of Machine Learning Research}, 3:1137--1155, 2003.

\bibitem{bengio1994learning}
Y.~Bengio, P.~Simard, and P.~Frasconi.
\newblock Learning long-term dependencies with gradient descent is difficult.
\newblock {\em IEEE Transactions on Neural Networks}, pages 157--166, 1994.

\bibitem{Theano10}
J.~Bergstra, O.~Breuleux, F.~Bastien, P.~Lamblin, R.~Pascanu, G.~Desjardins,
  J.~Turian, D.~Warde-Farley, and Y.~Bengio.
\newblock Theano: a {CPU} and {GPU} math expression compiler.
\newblock In {\em In Proc. of SciPy}, 2010.

\bibitem{Boldi2008}
P.~Boldi, F.~Bonchi, C.~Castillo, D.~Donato, A.~Gionis, and S.~Vigna.
\newblock The query-flow graph: Model and applications.
\newblock In {\em In Proc. of CIKM}, pages 609--618, 2008.

\bibitem{Bonchi2012}
F.~Bonchi, R.~Perego, F.~Silvestri, H.~Vahabi, and R.~Venturini.
\newblock Efficient query recommendations in the long tail via center-piece
  subgraphs.
\newblock In {\em In Proc. of SIGIR}, pages 345--354, 2012.

\bibitem{Broccolo2012}
D.~Broccolo, L.~Marcon, F.~M. Nardini, R.~Perego, and F.~Silvestri.
\newblock Generating suggestions for queries in the long tail with an inverted
  index.
\newblock {\em Inf. Process. Manage.}, 48(2):326--339, 2012.

\bibitem{Cao2008}
H.~Cao, D.~Jiang, J.~Pei, Q.~He, Z.~Liao, E.~Chen, and H.~Li.
\newblock Context-aware query suggestion by mining click-through and session
  data.
\newblock In {\em In Proc. of SIGKDD}, pages 875--883, 2008.

\bibitem{Kyung14}
K.~Cho, B.~Merrienboer, {\c{C}}.~G{\"{u}}l{\c{c}}ehre, F.~Bougares, H.~Schwenk,
  and Y.~Bengio.
\newblock Learning phrase representations using rnn encoder-decoder for
  statistical machine translation.
\newblock {\em In Proc. of EMNLP}, 2014.

\bibitem{clarke2011overview}
C.~LA Clarke, N.~Craswell, I.~Soboroff, and E.~M Voorhees.
\newblock Overview of the trec 2011 web track.
\newblock {\em Proceedings of the 2011 Text Retrieval Conference (TREC 2011)},
  2011.

\bibitem{Graves13}
A.~Graves.
\newblock Generating sequences with recurrent neural networks.
\newblock {\em CoRR}, abs/1308.0850, 2013.

\bibitem{He2009}
Q.~He, D.~Jiang, Z.~Liao, S.~C.~H. Hoi, K.~Chang, E.~P. Lim, and H.~Li.
\newblock Web query recommendation via sequential query prediction.
\newblock In {\em In Proc. of ICDE}, pages 1443--1454, 2009.

\bibitem{El1995hierarchical}
S.~El Hihi and Y.~Bengio.
\newblock Hierarchical recurrent neural networks for long-term dependencies.
\newblock In {\em NIPS}, pages 493--499. Citeseer, 1995.

\bibitem{hochreiter1997long}
S.~Hochreiter and J.~Schmidhuber.
\newblock Long short-term memory.
\newblock {\em Neural computation}, 9(8):1735--1780, 1997.

\bibitem{Huang2009}
J.~Huang and E.~N. Efthimiadis.
\newblock Analyzing and evaluating query reformulation strategies in web search
  logs.
\newblock In {\em In Proc. of CIKM}, pages 77--86, 2009.

\bibitem{huang2013learning}
P.~S. Huang, X.~He, J.~Gao, L.~Deng, A.~Acero, and L.~Heck.
\newblock Learning deep structured semantic models for web search using
  clickthrough data.
\newblock In {\em In Proc. of CIKM}, pages 2333--2338, 2013.

\bibitem{Jain2011}
A.~Jain, U.~Ozertem, and E.~Velipasaoglu.
\newblock Synthesizing high utility suggestions for rare web search queries.
\newblock In {\em In Proc. of SIGIR}, pages 805--814, 2011.

\bibitem{Jansen2007}
B.~J. Jansen, A.~Spink, C.~Blakely, and S.~Koshman.
\newblock Defining a session on web search engines: Research articles.
\newblock {\em J. Am. Soc. Inf. Sci. Technol.}, 58(6):862--871, April 2007.

\bibitem{Jiang2014}
J.Y. Jiang, Y.Y. Ke, P.Y. Chien, and P.J. Cheng.
\newblock Learning user reformulation behavior for query auto-completion.
\newblock In {\em In Proc. of SIGIR}, pages 445--454, 2014.

\bibitem{Koehn2009}
P.~Koehn.
\newblock {\em Statistical machine translation}.
\newblock Cambridge University Press, 2009.

\bibitem{lideep}
X.~Li, C.~Guo, W.~Chu, Y.Y. Wang, and J.~Shavlik.
\newblock Deep learning powered in-session contextual ranking using
  clickthrough data.
\newblock In {\em In Proc. of NIPS}, 2014.

\bibitem{mikolov2013efficient}
T.~Mikolov, K.~Chen, G.~Corrado, and J.~Dean.
\newblock Efficient estimation of word representations in vector space.
\newblock {\em arXiv preprint arXiv:1301.3781}, 2013.

\bibitem{Mikolov2010}
T.~Mikolov, M.~Karafi{\'a}t, L.~Burget, J.~Cernock{\'y}, and S.~Khudanpur.
\newblock {Recurrent neural network based language model}.
\newblock In {\em In Proc. of ACISCA}, pages 1045--1048, 2010.

\bibitem{Bhaskar15}
B.~Mitra.
\newblock Exploring session context using distributed representations of
  queries and reformulations.
\newblock In {\em In Proc. of SIGIR}. To appear, 2015.

\bibitem{Ozertem2012}
U.~Ozertem, O.~Chapelle, P.~Donmez, and E.~Velipasaoglu.
\newblock Learning to suggest: A machine learning framework for ranking query
  suggestions.
\newblock In {\em In Proc. of SIGIR}, pages 25--34, 2012.

\bibitem{PascanuDeep13}
R.~Pascanu, C.~Gulcehre, K.~Cho, and Y.~Bengio.
\newblock How to construct deep recurrent neural networks.
\newblock {\em arXiv preprint arXiv:1312.6026}, 2013.

\bibitem{Pascanu13}
R.~Pascanu, T.~Mikolov, and Y.~Bengio.
\newblock On the difficulty of training recurrent neural networks.
\newblock {\em In Proc. of ICML}, 2013.

\bibitem{Raman2014}
K.~Raman, P.~Bennett, and K.~Collins-Thompson.
\newblock Understanding intrinsic diversity in web search: Improving
  whole-session relevance.
\newblock {\em ACM Trans. Inf. Syst.}, 32(4):20:1--20:45, October 2014.

\bibitem{BPTT1988}
D.~E Rumelhart, G.~E Hinton, and R.~J Williams.
\newblock Learning representations by back-propagating errors.
\newblock {\em Cognitive modeling}, 1988.

\bibitem{Sadikov2010}
E.~Sadikov, J.~Madhavan, L.~Wang, and A.~Halevy.
\newblock Clustering query refinements by user intent.
\newblock In {\em In Proc. of WWW}, pages 841--850, 2010.

\bibitem{Santos2013}
R.~L.T. Santos, C.~Macdonald, and I.~Ounis.
\newblock Learning to rank query suggestions for adhoc and diversity search.
\newblock {\em Information Retrieval}, 16(4):429--451, 2013.

\bibitem{shen2014latent}
Y.~Shen, X.~He, J.~Gao, L.~Deng, and G.~Mesnil.
\newblock A latent semantic model with convolutional-pooling structure for
  information retrieval.
\newblock In {\em In Proc. of CIKM}, pages 101--110, 2014.

\bibitem{Shok2013}
M.~Shokouhi.
\newblock Learning to personalize query auto-completion.
\newblock In {\em In Proc. of SIGIR}, pages 103--112, 2013.

\bibitem{sordoni2013modeling}
A.~Sordoni, J.Y. Nie, and Y.~Bengio.
\newblock Modeling term dependencies with quantum language models for ir.
\newblock In {\em In Proc. of SIGIR}, pages 653--662, 2013.

\bibitem{Ilya14}
I.~Sutskever, O.~Vinyals, and Q.~V.~V Le.
\newblock Sequence to sequence learning with neural networks.
\newblock In {\em In Proc. of NIPS}, pages 3104--3112. 2014.

\bibitem{Templates11}
I.~Szpektor, A.~Gionis, and Y.~Maarek.
\newblock Improving recommendation for long-tail queries via templates.
\newblock In {\em In Proc. of WWW}, pages 47--56, 2011.

\bibitem{Vahabi2013}
H.~Vahabi, M.~Ackerman, D.~Loker, R.~Baeza-Yates, and A.~Lopez-Ortiz.
\newblock Orthogonal query recommendation.
\newblock In {\em In Proc. of RECSYS}, RecSys '13, pages 33--40. ACM, 2013.

\bibitem{Wen2001}
J.R. Wen, J.Y. Nie, and H.J. Zhang.
\newblock Clustering user queries of a search engine.
\newblock In {\em In Proc. of WWW}, pages 162--168. ACM, 2001.

\bibitem{mart}
Q.~Wu, C.~J. Burges, K.~M. Svore, and J.~Gao.
\newblock Adapting boosting for information retrieval measures.
\newblock {\em Inf. Retr.}, 13(3):254--270, June 2010.

\end{thebibliography}
\end{document}